\documentclass[10pt,twocolumn,letterpaper]{article}

\usepackage{cvpr}            %

\definecolor{cvprblue}{rgb}{0.21,0.49,0.74}
\usepackage[pagebackref,breaklinks,colorlinks,allcolors=cvprblue]{hyperref}
\usepackage{url}
\usepackage{booktabs}
\usepackage{multirow}
\usepackage{adjustbox}
\usepackage{wrapfig}
\usepackage{mathtools}
\usepackage[table]{xcolor}
\usepackage{bm}
\usepackage[normalem]{ulem}
\usepackage{tcolorbox}
\usepackage{float}
\usepackage{cuted}
\tcbuselibrary{listings, breakable, skins}
\definecolor{MyBlue}{RGB}{98, 193, 229}

\newcommand{\method}{$\text{R}^2$VLM\xspace}
\newcommand{\methodzero}{$\text{R}^2$VLM-Zero\xspace}
\newcommand{\methodsft}{$\text{R}^2$VLM-SFT\xspace}

\def\paperID{***} %
\def\confName{CVPR}
\def\confYear{2026}

\title{Recurrent Reasoning with Vision-Language Models for \\Estimating Long-Horizon Embodied Task Progress}

\author{
Yuelin Zhang$^{1,2,3,4}$\thanks{Equal contribution, $^{\dag}$ Corresponding author.}~\qquad
Sijie Cheng$^{4,5,6}$\footnotemark[1]~\qquad
Chen Li$^{1,2,3}$~\qquad
Zongzhao Li$^{1,2,3}$~\qquad
Yuxin Huang$^{1,2,3}$~\qquad \\
Yang Liu$^{5,6}$~\qquad
Wenbing Huang$^{1,2,3~\dag}$~\qquad\\
$^1$~Gaoling School of Artificial Intelligence, Renmin University of China\\
$^2$~Beijing Key Laboratory of Research on Large Models and Intelligent Governance\\
$^3$~Engineering Research Center of Next-Generation Intelligent Search and Recommendation, MOE\\
$^4$~RayNeo.AI
$^5$~Department of Computer Science and Technology, Tsinghua University\\
$^6$~Institute for AI Industry Research (AIR), Tsinghua University\\
}

\begin{document}
\maketitle

\begin{abstract}

Accurately estimating task progress is critical for embodied agents to plan and execute long-horizon, multi-step tasks. 
Despite promising advances, existing Vision-Language Models (VLMs) based methods primarily leverage their video understanding capabilities, while neglecting their complex reasoning potential. Furthermore, processing long video trajectories with VLMs is computationally prohibitive for real-world deployment. To address these challenges, we propose the Recurrent Reasoning Vision-Language Model (\method). Our model features a recurrent reasoning framework that processes local video snippets iteratively, maintaining a global context through an evolving Chain of Thought (CoT). This CoT explicitly records task decomposition, key steps, and their completion status, enabling the model to reason about complex temporal dependencies. This design avoids the high cost of processing long videos while preserving essential reasoning capabilities. We train \method on large-scale, automatically generated datasets from ALFRED and Ego4D. Extensive experiments on progress estimation and downstream applications, including progress-enhanced policy learning, reward modeling for reinforcement learning, and proactive assistance, demonstrate that \method achieves strong performance and generalization, achieving a new state-of-the-art in long-horizon task progress estimation. The models and benchmarks are publicly available at \href{https://huggingface.co/collections/zhangyuelin/r2vlm}{huggingface}.

\end{abstract}

\section{Introduction}

Accurately estimating task progress is essential for embodied agents in performing complex, multi-step tasks. By observing and recording actions, agents must assess their advancement toward the goal, including what has been completed and what remains. This internal tracking of progress enables long-horizon planning and context-aware decision-making. For instance, to address the query ``Please make me a cup of coffee", an agent can proceed to the subtask \textit{navigate\_to\_coffee\_machine} if \textit{pick\_cup} is 100\% complete; and it can trigger a recovery behavior if progress stalls.

\begin{figure}[t!]
    \centering
    \includegraphics[width=0.98\linewidth]{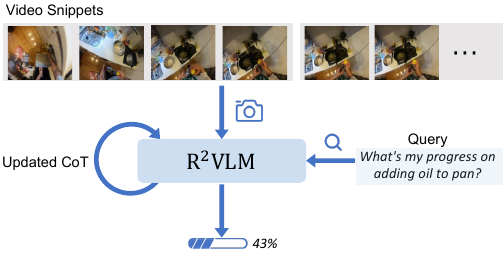}
    \caption{\method performs multi-turn recurrent reasoning. It takes a video snippet and the CoT from the previous iteration as input, and outputs the updated CoT and a progress estimation.}  %
    \label{fig:titlefigure}
    \vskip -0.2in
\end{figure}

Recently, GVL~\cite{du2023vision} and ROVER~\cite{schroeder2025rover} have explored using Vision-Language Models (VLMs) for task progress estimation. However, their approaches primarily leverage VLMs' video understanding capabilities and large context windows, leaving their reasoning potential underexplored. Accurately estimating progress in long-horizon, egocentric tasks remains highly challenging for two main reasons. First, embodied tasks often consist of multiple subtasks with complex temporal dependencies. A model must therefore reason about these dependencies and align them with visual observations. Second, the long video trajectories involved incur substantial computational and time costs when processed by VLMs, rendering such approaches impractical for real-world deployment. Addressing these challenges is essential to developing a progress estimation model that is stable, accurate, and practical for real-world use.

\begin{figure*}[t!]
    \centering
    \includegraphics[width=0.9\linewidth]{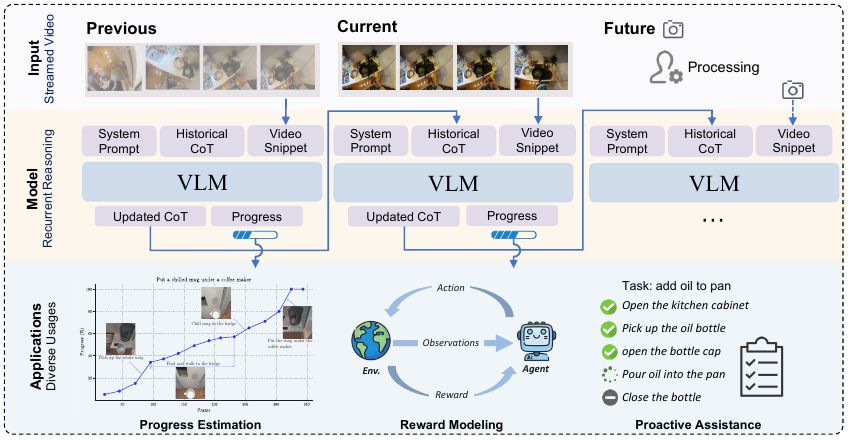}
    \caption{Framework of \method. It performs recurrent reasoning over the newly observed video snippet, while incorporating the historical CoT as context. The CoT records the completed, pending, and incomplete steps from earlier video snippets and is dynamically updated throughout the recurrent reasoning process. The model estimates task progress based on the proportion of completed steps, and supports diverse applications including progress estimation, reward modeling, and proactive assistance.}
    \vskip -0.1in
    \label{fig:model_framework}
\end{figure*}

In this work, we introduce the \textbf{R}ecurrent \textbf{R}easoning \textbf{V}ision-\textbf{L}anguage \textbf{M}odel (\method) for long-horizon embodied task progress estimation. As illustrated in \cref{fig:model_framework}, \method introduces two key innovations to address the aforementioned challenges.
First, analogous to standard reasoning VLMs, we endow \method with a Chain of Thought (CoT) that precedes its final progress estimation, which equips the model with the reasoning capability needed to handle complex temporal dependencies between subtasks. Our CoT is specifically designed to record the high-level task decomposition, the required key steps, and their completion status.
Second, unlike standard VLMs, our model performs recurrent reasoning over local video snippets rather than the full, long video. At each temporal iteration, \method updates its CoT by integrating the historical CoT with the currently observed video snippet. This strategy provides two major benefits: it avoids the redundant overhead of processing a large number of visual tokens, and it maintains crucial global context through the evolving CoT, even when visual inputs are locally restricted.

To evaluate the progress estimation capability in long-horizon embodied tasks, we conduct experiments on both the simulated ALFRED dataset~\cite{shridhar2020alfred} and real-world Ego4D dataset~\cite{grauman2022ego4d}.
Specifically, we develop an automated data generation pipeline to build a large-scale, multi-turn incremental video interaction dataset for training and evaluating.
Experimental results demonstrate that \method achieves superior performance across the two benchmarks and four metrics, validating the effectiveness of both recurrent reasoning and specifically designed rewards.
Furthermore, we extensively explore three potential downstream applications, including progress-enhanced embodied policy learning, reward modeling for online reinforcement learning, and step-wise proactive assistance. 
Extensive experiments confirm that \method exhibits strong generalization ability and promising application potential for embodied AI research.

\section{Related Work}

\textbf{Large Pre-trained Models as Progress Estimator.}
In embodied and robotic tasks, many prior studies have explored using large models to give progress estimation. Early methods~\cite{ma2022vip,ma2023liv} fine-tune pretrained image or text encoders and compute the feature distance between the current observation and the goal state to represent progress. However, these methods operate primarily at the image level and fail to capture temporal processes, leading to degraded performance in long-horizon tasks. %
Many video-based works leverage the context window of VLMs to naturally maintain global information. GVL~\cite{du2023vision} and ROVER~\cite{schroeder2025rover} are two methods based on in-context learning, which design innovative prompting strategies or inference trajectories to enhance the VLMs' progress estimation capability.
However, as these methods fully depend on pretrained VLMs and overlook the VLMs' reasoning capabilities, their performance remains limited on complex long-horizon tasks. Although~\citet{du2023vision} fine-tunes Flamingo as a success detector, it only provides sparse binary feedback. In contrast, our model integrates temporal modeling of long-horizon multi-step execution with reinforcement learning, enabling stronger reasoning and delivering dense, accurate progress estimation.

\noindent\textbf{Reward Modeling for Long-Horizon Tasks.}
Early reward learning often uses inverse reinforcement learning (IRL;~\cite{ng2000algorithms,kumar2023graph,krishnan2016hirl}), which infers latent rewards from expert demonstrations but is unstable under noise and hard to scale to long horizons.
For long-horizon manipulation tasks, many studies~\cite{hung2024victor,chen2025sarm} have proposed hierarchical reward frameworks. These methods decompose complex tasks into multi-level subtasks according to the semantics and sequentially estimate the reward at different levels.
In contrast, our model supports end-to-end reward modeling, directly using progress changes inferred from visual observations as rewards. Without manual hierarchical decomposition, it learns task decomposition and reasoning strategies inherently through training. This enables smooth, context-aware reward signals for long-horizon tasks while maintaining interpretability and robustness.

\section{\method}

In this section, we introduce the core design and training strategy of \method. We begin by formally defining the progress estimation task in~\cref{sec:task_formulation}. Then, we present the proposed recurrent reasoning framework in~\cref{sec:recurrent-reasoning}. In~\cref {sec:dataset}, we describe the automated dataset construction pipeline. Finally, in~\cref{sec:RL-training}, we introduce the two-stage training procedure and the design of the RL rewards.

\subsection{Task Formulation}
\label{sec:task_formulation}
\cref{fig:model_framework} presents the overall framework of \method. At the input stage, we employ a streamed video input scheme, where the model continuously receives egocentric video of an agent performing tasks and periodically feeds the accumulated video snippet into the model for inference.
Using video snippet inputs enables the model to estimate the operator's progress in near real time, while avoiding the redundant computation and delay that would result from reprocessing the full video at every inference step. 

Let $v$ denote the egocentric video of an agent performing a task query $\tau$. At each iteration $t$, the goal is to estimate the normalized task progress $p_t \in [0,100]$ given only the visual observations.
To reduce the inference latency, the video is processed as a sequence of short, local snippets.
\begin{align}
    v = \{ v_t\}_{t=1}^T, \quad 
    v_t = \{I_t^{(k)}\}_{k=1}^K,
\end{align}
where each video snippet $v_t$ contains $K$ uniformly sampled image frames $I_t^{(k)}$. Given the task description $\tau$ and current video snippet $v_t$, the goal of task progress estimation is to learn a parametric VLM $f_\theta$ that predicts $p_t$:
\begin{align}
\label{eq:task}
   p_t = f_\theta(\tau,v_t). 
\end{align}

Our model, being reasoning-based, produces a CoT $c_t$ before generating the final progress estimate $p_t$. Crucially, we employ a recurrent framework: at each iteration $t$, the model $f_\theta$ takes the historical CoT $c_{t-1}$ and a current video snippet $v_t$ as input, incrementally updates the CoT to $c_t$, and then outputs the current progress value $p_t$. As the CoT explicitly encodes information from previous video snippets, it can serve as a form of historical memory.
Formally, we recast the formulation in~\cref{eq:task} as the recurrent form:
\begin{align}
    c_t, p_t = f_\theta(\tau, v_t, c_{t-1}),
\end{align}
where each token in the CoT $c_t$ is predicted autoregressively, and the final progress estimate $p_t$ is conditioned on this generated chain of thought. 

In what follows, we will first introduce how to recurrently update $c_{t-1}$ to $c_t$ in~\cref{sec:recurrent-reasoning}, and then present how to train the model $f_\theta$ via reinforcement learning in~\cref{sec:RL-training}.

\subsection{Recurrent Reasoning}
\label{sec:recurrent-reasoning}

\begin{figure}[t!]
    \centering
    \includegraphics[width=0.98\linewidth]{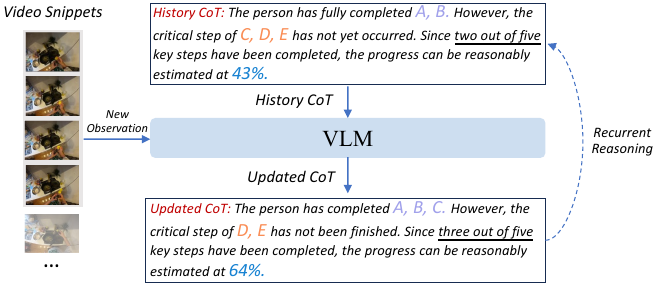}
    \caption{An illustration of the CoT structure and its update process in recurrent reasoning.} 
    \label{fig:cotdemo}
    \vskip -0.15in
\end{figure}

At the initial iteration, we prompt the VLM to generate an initial CoT $c_0$ using its inherent commonsense knowledge. For a multi-step task described in natural language, an ideal CoT should comprise three sequential elements: (i) a decomposition of the long-horizon task, (ii) an analysis of which key subtasks have been observed and which have not, and (iii) a progress estimate derived from the proportion of completed subtasks.
We explicitly specify these reasoning requirements in the system prompt (\cref{sec:appendix_model_system_prompt}), guiding the model to generate a well-structured CoT.

Since the environment is only partially observed at each iteration, the decomposition in $c_{t-1}$ may not fully align with the agent's actual execution. Therefore, during subsequent reasoning iterations, the model can dynamically adjust the task decomposition by merging, splitting, or reordering steps. It then updates the completion status in $c_{t}$ based on the new video snippet $v_t$. The main structure and update procedure of the CoT are illustrated in \cref{fig:cotdemo}.

Our recurrent reasoning framework offers three key advantages. First, by explicitly generating a CoT prior to the final output, the model enhances both the accuracy and interpretability of its progress estimations. Second, the recurrent integration of the updated CoT supplies essential historical context to the local video snippet, thereby preserving global task coherence. Finally, and most importantly, this incorporation of historical reasoning allows the model to inherit and refine prior task decompositions, ensuring logical consistency and stability across multi-step estimations.

\subsection{Dataset Construction}
\label{sec:dataset}

As previous studies seldom explore progress estimation with CoT guidance, we build an automated pipeline that converts expert demonstrations from public datasets into video snippets paired with corresponding CoTs. 
We utilize expert trajectories from ALFRED~\cite{shridhar2020alfred} (simulation) and Ego4D~\cite{grauman2022ego4d} (real world), both datasets can be expressed in a unified format: $\{\mathcal{T}, \mathcal{E}, (s_i, t^s_i, t^e_i), i=1\dots n\}$, where $\mathcal{T}$ is the task description, $\mathcal{E}$ the video demonstration,  each step $s_i$ is paired with its start and end timestamps, $t^s_i$ and $t^e_i$. We additionally construct a benchmark of unseen scenarios, which is manually reviewed, with an 83\% retention rate. More details are provided in \cref{sec:appendix_dataset_statistics}.

\noindent\textbf{Construct Video Snippets with Progress Labels.} 
We divide the expert video $\mathcal{E}$ into short snippets (4s for ALFRED and 2s for Ego4D) and uniformly sample 4 frames from each snippet, resulting in input frame rates of 1 fps and 2 fps, respectively. Such downsampling rarely loses crucial information.
We define progress as the proportion of completed steps, and assume linear variation within each step. This step-based formulation better reflects long-horizon task structure than time-based linearization. Given the timestamp $t$ of the last input frame, the progress is:
\begin{align}
    \textstyle  p_t \;=\; 100 \cdot (\frac{k}{n} \;+\; \frac{1}{n}\cdot \frac{t - t_{k+1}^s}{t_{k+1}^e - t_{k+1}^s}), t \in [t_{k+1}^s, \, t_{k+1}^e],
\end{align}
where $k$ denotes the maximum number of fully completed steps at timestep $t$.

\noindent\textbf{Generate Distractor Task Descriptions.}
Training solely on expert demonstrations may cause the model to rely on superficial visual cues as shortcuts, rather than truly reasoning over the task structure and step logic. To improve robustness, we generate distractor task descriptions leveraging powerful VLMs.
Note that directly prompting VLM to generate unrelated descriptions is infeasible as they lack definable progress. We propose a constrained prompting strategy that forces distractor tasks to match early steps but diverge from later ones; details are provided in~\cref{sec:appendix_distractor_detail}. This allows controllable progress assignment. Assuming the former $n_r$ steps are relevant to the distractor task, the progress is defined as:
\begin{align}
    \textstyle p_t \;=\; 100 \cdot \min\!\left( \frac{n_r}{n}, \; \frac{k}{n} + \frac{1}{n}\cdot \frac{t - t_{k+1}^s}{t_{k+1}^e - t_{k+1}^s} \right).
\end{align}

\noindent\textbf{Generate Chain-of-Thought Training Data.} 
We find that directly applying RL to VLMs fails to elicit fine-grained progress analysis in long-horizon tasks. Moreover, as CoT serves as contextual input for subsequent inference in our framework, it imposes higher demands on the quality of the reasoning chain.
To address this, we develop an automated pipeline that leverages large models to integrate prior knowledge and generate high-quality CoT data.
As mentioned before, an ideal CoT should sequentially decompose the task, analyze observed versus unobserved key steps in the video, and estimate task progress from the proportion of completed steps. Notably, the above critical information required to construct such a CoT is already available as prior knowledge in the dataset. Thus, we can directly provide it to an LLM for integration, making high-quality CoT generation straightforward. The prompting template and sample outputs are provided in \cref{sec:appendix_cot_generation_prompt}.

\noindent\textbf{Dataset Statistics.}
For the datasets, we constructed 11,499 ALFRED trajectories containing 124,821 dialogue tuples (\emph{i.e.}, $(v_t, c_t, p_t)$), and 13,965 Ego4D trajectories with 127,694 dialogue tuples.
For the benchmarks, three volunteers manually reviewed the generated data. In ALFRED, 669 of the 722 automatically annotated samples were retained, resulting in a 93\% retention rate. In Ego4D, 529 of the 718 samples were kept, yielding a 74\% retention rate.

\subsection{Training Strategy}
\label{sec:RL-training}

Leveraging the generated CoT data, the training of $f_\theta$ consists of two stages: Supervised Fine-tuning (SFT) and multi-turn Reinforcement Learning (RL). 
We first perform SFT to help the model learn reasoning patterns for task decomposition, step-wise reasoning, and progress estimation. Based on an early checkpoint from the SFT stage as the cold-start checkpoint, we further apply RL to improve the model's performance on the multi-step reasoning task.

In our implementation, we use Proximal Policy Optimization (PPO;~\cite{schulman2017proximal}) instead of Group Relative Policy Optimization (GRPO;~\cite{shao2024deepseekmath}), because PPO computes advantages independently for each sample, whereas GRPO requires multiple candidates generated from an identical input, which is incompatible with our multi-turn setting where the input $c_{t-1}$ differs across trajectories. We initialize the actor, critic, and reference models all from the cold-start checkpoint. 
At iteration $t$, let $p_t$ and $p_t^{\text{gt}}$ denote the predicted and ground-truth progress, respectively, while $n^{\text{gt}}$ and $m_t^{\text{gt}}$ represent the ground-truth total number of subtasks and the number of completed subtasks.
The reward functions are formulated as follows:

\noindent\textbf{Format Reward.} The Format Reward checks whether the model's output follows the required format of \textless think\textgreater reasoning process\textless /think\textgreater\textless answer\textgreater progress estimation\textless /answer\textgreater. If in valid format, $R_\text{fmt} = 1$; otherwise, $R_\text{fmt} = 0$.

\noindent\textbf{Bin Reward.} For long-horizon tasks consisting of multiple steps (\emph{i.e.} subtasks), we divide the overall progress range into several discrete bins according to the number of steps.
The Bin Reward requires the model's predicted progress to fall within the corresponding interval. This reward not only provides a coarse-grained constraint on prediction accuracy but also reflects the model's ability to assess stage-wise completion status. It is defined as follows:
\begin{align}
    R_\text{bin} =
    \textstyle
    \begin{cases}
        1.0,  & \text{if } \frac{m_t^\text{gt}}{n^\text{gt}} \leq \frac{p_t}{100} <\frac{ (m_t^\text{gt} + 1)}{n^\text{gt}};  \\
        0.25, & \text{if } \frac{(m_t^\text{gt} - 1)}{n^\text{gt}} \leq \frac{p_t}{100} <\frac{ (m_t^\text{gt} - 1 + 1)}{n^\text{gt}};  \\
        0.25, & \text{if } \frac{(m_t^\text{gt} + 1)}{n^\text{gt}} \leq \frac{p_t}{100} <\frac{ (m_t^\text{gt} + 1 + 1)}{n^\text{gt}};  \\
        0.0.  & \text{otherwise}.
    \end{cases}
\end{align}
This means that if the model's estimated progress falls within the correct interval, the reward is set to 1; if it falls within an adjacent interval, the reward is 0.25; otherwise, the reward is 0.

\noindent\textbf{MAE Reward.} The MAE Reward encourages more accurate progress predictions, serving as a fine-grained constraint on the model's progress estimation. The formula is:
\begin{align}
    \textstyle R_\text{mae} = \text{max}(1 - \lvert p_t-p_t^\text{gt}\rvert/\delta_1, 0),
\end{align}
where $\delta_1$ is a hyperparameter controlling the tolerance between the predicted and ground-truth progress.

\noindent\textbf{Improvement Reward.} Specifically designed for multi-turn RL, we introduce the Improvement Reward that encourages the new iteration to produce smaller error than the previous one. Because each iteration conditions on the CoT generated in the previous turn as part of its contextual input, generating an improved output in the new iteration directly reflects the model’s capacity for recurrent reasoning and self-correction. The reward is therefore computed as:
\begin{align}
    \textstyle R_{\text{imp}} &= (\lvert p_{t-1}-p_{t-1}^{\text{gt}} \rvert - \lvert p_t - p_t^{\text{gt}} \rvert) / \delta_2,
\end{align}
where $R_{\text{imp}}$ is derived from the error difference between two consecutive turns, and $\delta_2$ is also a hyperparameter. We normalize $R_{\text{imp}}$ to the asymmetric range [-1, 0.8] to amplify the penalty for increasing errors.

\noindent\textbf{Finish Reward.} Finish Reward is designed to evaluate the model's ability to identify full task completion, enforcing the generation of more accurate task completion signals.
\begin{align}
    \textstyle R_\text{fin}=
    \begin{cases}
      1.0, & \text{if } p_t = 100 \land p_t^\text{gt} = 100;  \\
      1.0, & \text{if } p_t \neq 100 \land p_t^\text{gt} \neq 100;  \\
      0.0, & \text{otherwise}.
    \end{cases}
\end{align}

\noindent\textbf{Overall Reward.} The overall reward is formulated as:
\begin{align}
    \textstyle R_\text{overall} = R_\text{fmt} \cdot (R_\text{bin} \cdot R_\text{mae} + \alpha R_\text{imp} + \beta R_\text{fin}),
\end{align}
where $\alpha$ and $\beta$ are hyperparameters. Here, the multiplicative term $R_\text{bin} \cdot R_\text{mae}$ ensures that the model receives fine-grained MAE reward only when the basic bin prediction is correct. This design helps improve training stability.

\section{Experiments}

\subsection{Experimental Setups}
\label{sec:experiment_setups}

\textbf{Implementation.} We train our model based on Qwen2.5-VL-7B-Instruct, using LLaMA-Factory~\cite{zheng2024llamafactory} for SFT and Verl~\cite{sheng2025hybridflow} for RL. %
In the main experiment, we conduct training and evaluation separately on ALFRED and Ego4D. The results of mixed training and cross-domain testing are presented in~\cref{sec:appendix_crossdomain}. Detailed training configurations are provided in~\cref{sec:appendix_training_details}.

\noindent\textbf{Baselines.} To comprehensively evaluate our model, we compare it with closed-source API-based models, %
advanced open-source models, %
and task-specific methods for progress estimation. %
All baselines operate in the standard setting, receiving the full video from the beginning up to the current time step. Due to the long duration of tasks and the model's limits on context length and image counts, we uniformly downsample the frames to 32, 16, etc., when the full video cannot be processed. Details on baseline models and their input frame counts are provided in~\cref{sec:appendix_baseline_model_information}.

\noindent\textbf{Metrics.} We design four metrics to comprehensively assess our model's performance, where each metric highlights a specific aspect of its practical applicability and corresponds to a distinct downstream task.
Detailed formulations of the four metrics are provided in~\cref{sec:appendix_evaluation_metric_formulations}.
\begin{itemize}
    \item $\bm{p}_{\textbf{mae}}$: the mean absolute error (MAE) between the predicted and actual progress, which reflects the model's capability in the progress estimation task.
    \item $\bm{\Delta p}_\textbf{mae}$: the MAE between the model's predicted and the ground-truth progress increment relative to the previous iteration.
    The increment in progress represents the immediate reward induced by the agent's action. This metric evaluates the model's capability in reward modeling.
    \item $\bm{bin}$: This metric checks whether the model's predicted progress falls within the correct bin, which reflects its ability to evaluate the completion status of individual steps in long-horizon multistep tasks. It aligns with the capability required by proactive assistance.
    \item $\bm{acc}$: This metric checks the correctness of the model's prediction of overall task completion, which reflects its ability to provide binary rewards for long-horizon tasks.
\end{itemize}

\subsection{Main Results}
\label{sec:main-results}

\begin{table*}[t!]
    \centering
    \setlength{\tabcolsep}{13pt}
    \caption{We evaluated our three model variants and all baselines on the Alfred and Ego4D benchmarks. \method outperforms every baseline on all four metrics across both datasets, reaching state-of-the-art performance.}
    \label{tab:main_results}
    \resizebox{\textwidth}{!} {
        \begin{tabular}{l|c|cccc|cccc}
        \toprule
        \multicolumn{1}{l}{\multirow{2}{*}{Model}} 
        & \multicolumn{1}{c}{\multirow{2}{*}{Size}} 
        & \multicolumn{4}{c}{Alfred} 
        & \multicolumn{4}{c}{Ego4D} \\

        \multicolumn{1}{c}{} 
        & \multicolumn{1}{c}{} 
        & \multicolumn{1}{c}{$p_\text{mae} \downarrow$} 
        & \multicolumn{1}{c}{$\Delta p_\text{mae} \downarrow$} 
        & \multicolumn{1}{c}{$bin \uparrow$} 
        & \multicolumn{1}{c}{$acc \uparrow$} 
        & \multicolumn{1}{c}{$p_\text{mae} \downarrow$} 
        & \multicolumn{1}{c}{$\Delta p_\text{mae} \downarrow$} 
        & \multicolumn{1}{c}{$bin \uparrow$} 
        & \multicolumn{1}{c}{$acc \uparrow$} \\
        
        \midrule
        \rowcolor{gray!10}
        \multicolumn{10}{c}{\textbf{\textit{API-Based Models}}} \\ 
        \midrule
        
        GPT-5~\cite{openai2025gpt5}                      & - & 18.35 & 15.48 & 0.505 & 0.901 & 25.04 & 15.89 & 0.259 & 0.749 \\
        Gemini-2.5-Pro~\cite{comanici2025gemini}         & - & 16.27 & 17.37 & 0.481 & 0.830 & 28.22 & 16.71 & 0.217 & 0.713 \\
        Qwen3-VL-Plus~\cite{qwen2025qwen3vlmax}          & - & 27.30 & 15.29 & 0.379 & 0.777 & 28.20 & 10.14 & 0.251 & 0.652 \\ 
        
        \midrule
        \rowcolor{gray!10}
        \multicolumn{10}{c}{\textbf{\textit{Open-Source Models}}} \\ 
        \midrule
        MiniCPM-V-2.6~\cite{yao2024minicpm}              & 8B  & 27.47 & 22.04 & 0.478 & 0.373 & 30.17 & 18.03 & 0.208 & 0.482 \\
        GLM-4.1V-Thinking~\cite{hong2025glm}             & 9B  & 22.67 & 17.47 & 0.329 & 0.548 & 30.45 & 8.21 & 0.197 & 0.488 \\
        InternVL3~\cite{zhu2025internvl3}                & 8B  & 36.09 & 13.56 & 0.311 & 0.621 & 29.43 & 16.80 & 0.187 & 0.511 \\
        InternVL3~\cite{zhu2025internvl3}                & 78B & 30.54 & 16.65 & 0.360 & 0.765 & 28.16 & 13.37 & 0.204 & 0.586 \\
        Qwen2.5-VL~\cite{bai2025qwen25vl}                & 7B  & 27.87 & 10.67 & 0.295 & 0.494 & 28.32 & 10.11 & 0.206 & 0.485 \\
        Qwen2.5-VL~\cite{bai2025qwen25vl}                & 72B & 24.88 & 18.68 & 0.342 & 0.781 & 26.88 & 11.11 & 0.254 & 0.624 \\
        \midrule
        
        \rowcolor{gray!10}
        \multicolumn{10}{c}{\textbf{\textit{Specific Methods}}} \\ 
        \midrule
        LIV~\cite{ma2023liv}                             & $<$1B  & 35.87 & 32.44 & 0.153 & 0.490 & 34.57 & 24.51 & 0.151 & 0.485 \\
        ROVER~\cite{schroeder2025rover}                  & -      & 29.43 & 11.52 & 0.308 & 0.575 & 39.83 & 6.48  & 0.174 & 0.548 \\
        GVL-SFT~\cite{ma2024vision}                      & 7B     & 6.21  & 8.39  & 0.830 & 0.930 & 26.80 & 13.26 & 0.255 & 0.665 \\
        
        \midrule
        \methodzero                             & 7B  & 7.21  & 7.78  & 0.742 & 0.490 & 23.58 & 10.97 & 0.256 & 0.676 \\
        \methodsft                              & 7B  & 2.77  & 2.27  & 0.876 & 0.961 & 23.03 & 3.98 & 0.266 & 0.685 \\
        \rowcolor{blue!7.5}
        \method                                  & 7B  & \textbf{2.19}  & \textbf{2.17}  & \textbf{0.917} & \textbf{0.988} & \textbf{19.25} & \textbf{3.88} & \textbf{0.318} & \textbf{0.761} \\

        \bottomrule
        \end{tabular}
    }
    \vskip -0.1in
\end{table*}

Our model includes three variants: \methodzero (direct RL on the base model), \methodsft (SFT on the full generated CoT dataset), and \method (RL continued from cold-start checkpoint of \methodsft). 
Experimental results are presented in~\cref{tab:main_results}.
\textbf{\textit{Firstly}}, \methodsft and \method achieve significant improvements over the pretrained Qwen2.5-VL-7B across all metrics on both datasets. 
For example, \method achieves a $p_\text{mae}$ of 2.19 on ALFRED, substantially better than Qwen2.5-VL-7B (27.87) and Qwen2.5-VL-72B (24.88).
Pretrained models without domain-specific fine-tuning exhibit poor zero-shot progress estimation performance, with limited gains from in-context learning or novel reasoning strategies.
\textbf{\textit{Secondly}}, from a training perspective, further applying RL after CoT-based SFT helps the model overcome supervised signal limitations and learn more generalizable, adaptive reasoning strategies. 
For instance, on ALFRED, \methodsft attains a high accuracy of 0.961, while \method reaches an exceptional 0.988. In contrast, \methodzero fails to develop effective reasoning patterns, leading to poor performance.
\textbf{\textit{Thirdly}}, all models, including baselines, perform markedly better on ALFRED than Ego4D.
Notably, \method achieves outstanding results on ALFRED with progress MAE less than 3\% and both step-level ($bin$) and task-level accuracy ($acc$) above 0.9, while performance declines on the real-world Ego4D benchmark. 
This is primarily attributed to two factors:
(i) more complex, noisy real-world environments with distracting elements;
(ii) varied execution paths of the same task across human operators, challenging the model’s adaptive reasoning and progress estimation.
In contrast, ALFRED’s cleaner, structured environment and fixed task sequences enable superior model performance.

\subsection{Ablation Studies}
\label{sec:ablation_studies}

\begin{figure}[t!]
    \centering
    \includegraphics[width=0.48\linewidth]{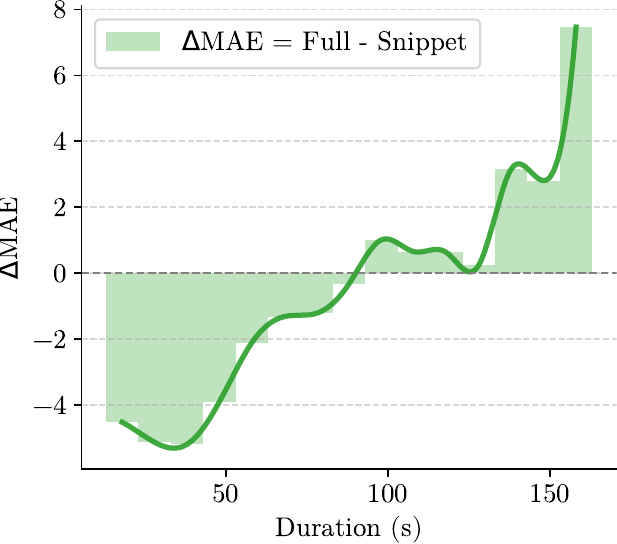}
    \hfill
    \includegraphics[width=0.48\linewidth]{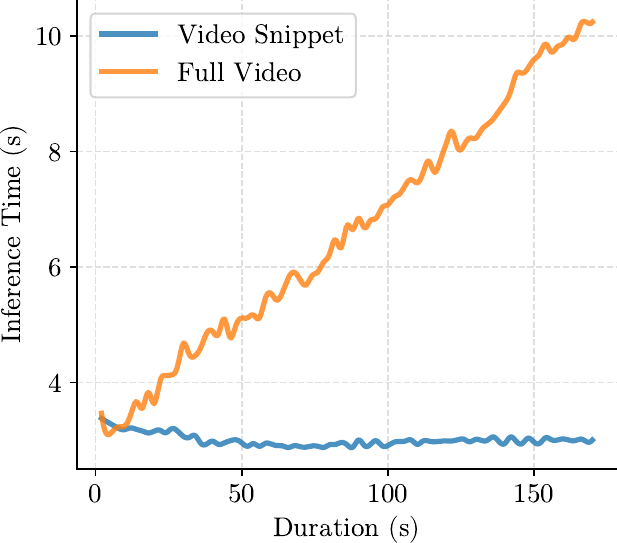}
    \caption{Comparison of MAE and inference time between full and incremental video inputs at different video durations.}
    \label{fig:increment_vs_full_video_input}
    \vskip -0.1in
\end{figure}

In this section, we perform ablation studies to validate the effectiveness of our core components, focusing on two key aspects: 
(1) the impact of using video snippet combined with historical CoT as input, compared to feeding the entire video sequence into the VLMs; and (2) the reward design for training \method in reinforcement learning.

\begin{figure*}[t!]
    \centering
    \includegraphics[width=\linewidth]{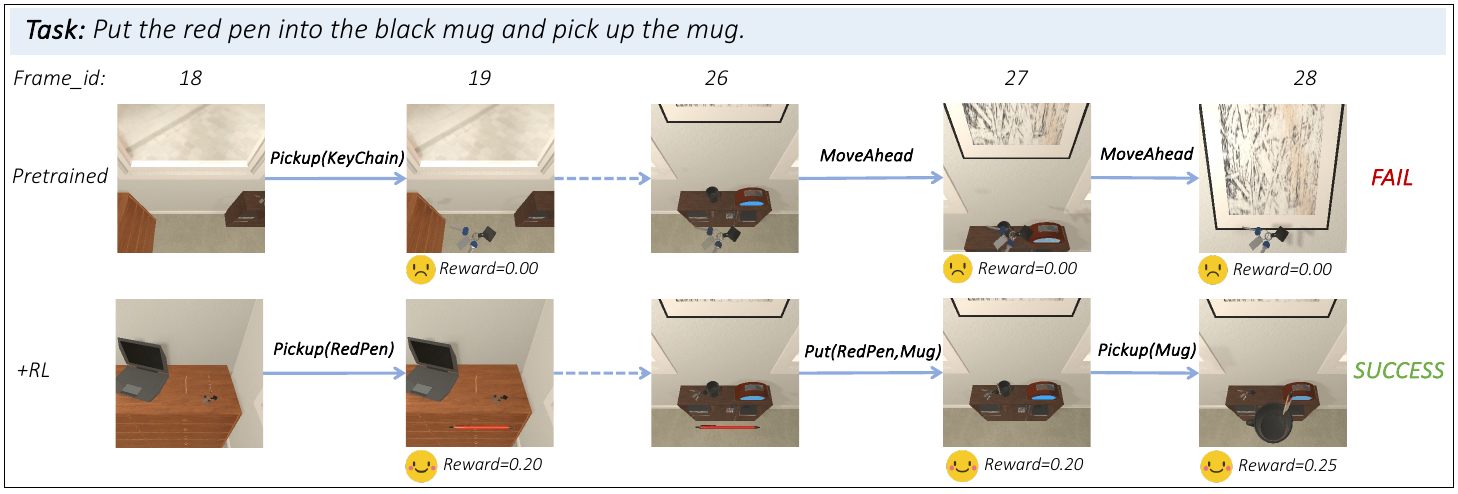}
    \caption{The comparison of the policy model's performance on a three-step task before and after online reinforcement learning. The rewards in the reinforcement learning process are provided by \method.} %
    \label{fig:reward_modeling}
\end{figure*}

\noindent\textbf{Video Snippet Input vs. Full Video Input.} We evaluate progress prediction accuracy and inference time on Ego4D by comparing full video input against video snippets combined with historical CoT.
Using an identical training setup, we fine-tune a variant with full video input, which shows poor training efficiency due to extremely long and variable context lengths that hinder batch parallelism. 
Results are presented in~\cref{fig:increment_vs_full_video_input}.
The left subfigure shows MAE differences across video durations: as execution time increases, the full-video-based method input yields progressively worse accuracy than the snippet-based one. This arises because VLMs processing overly long videos suffer from attention dispersion and hallucination, degrading predictions.
The right subfigure compares inference time: our snippet-based model maintains constant response time regardless of video length, while full video input leads to nearly linear growth in response time due to escalating computational costs.
For real-time online progress estimation in long-horizon tasks, using video snippets with historical CoT ensures high efficiency, alleviates long-sequence attention distraction, and improves prediction stability/accuracy.

\begin{table}[t!]
    \centering
    \caption{Ablation on the reward design for training \method with reinforcement learning.}
    \label{tab:ablation_reward_design}
    \setlength{\tabcolsep}{12pt}
    \begin{adjustbox}{width=\linewidth}
        \begin{tabular}{lcccc}
          \toprule
          & $p_\text{mae}\downarrow$ & $\Delta p_\text{mae}\downarrow$ & $bin\uparrow$ & $acc\uparrow$ \\
          \midrule
            w/o Bin Reward    & 2.22 & 2.26 & 0.912 & 0.984 \\
            w/o MAE Reward    & 2.22 & 2.23 & \textbf{0.924} & 0.981 \\
            w/o Improvement Reward  & 2.42 & 2.43 & 0.920 & 0.978 \\
            Ours              & \textbf{2.19} & \textbf{2.17} & 0.917 & \textbf{0.988} \\
          \bottomrule
        \end{tabular}
    \end{adjustbox}
    \vskip -0.1in
\end{table}

\noindent\textbf{Reward Design.}
\label{sec:ablation_reward_design}
We ablate three critical reward components: bin, MAE, and improvement rewards. 
In each experiment, one reward is removed while the others are retained using the same configurations. 
The results on the ALFRED benchmark are presented in~\cref{tab:ablation_reward_design}.
The model achieves optimal performance when all three rewards are combined.
Removing the MAE reward yields a slight gain in bin metrics but significantly degrades progress estimation ($p_\text{mae}$, $\Delta p_\text{mae}$) and task completion accuracy.
The improvement reward, defined by progress estimation accuracy gained over the previous turn, is the most essential: its removal increases $p_\text{mae}$ from 2.19 to 2.42 ($\approx$10.5\% error rise) and reduces $acc$ from 0.988 to 0.978. 
This confirms the improvement reward effectively guides iterative reasoning refinement, enabling continuous error correction and enhanced prediction accuracy in long-horizon tasks.

\subsection{Downstream Applications}
\label{sec:application}

\textbf{Progress-enhanced Embodied Policy Learning.}
Our model can serve as a general progress estimator for embodied tasks. We evaluate its effectiveness using the Seq2Seq model with progress monitoring from ALFRED~\cite{shridhar2020alfred}, where progress supervision was introduced to help agents assess state utility and improve performance. Most datasets lack explicit progress labels, so prior work relies on time-based progress, an approximation that often fails to reflect true task advancement. In contrast, our model learns progress from the proportion of completed steps, providing a more faithful and semantically grounded signal. We train Seq2Seq models with either time-based progress or progress estimated by \method and compare their goal and task completion rates in~\cref{tab:application_progress_estimation}.
Results show that time-based progress even degrades performance on the valid\_seen split. In contrast, using \method estimated progress yields clear gains. The goal-condition success improves by 2.2\% on valid\_seen and 0.4\% on valid\_unseen. These findings demonstrate both the practical value of our progress estimator for policy learning and the advantage of defining progress via completed steps rather than time.

\newcommand{\B}[1]{\textbf{#1}}

\begin{table}[t!]
  \centering
    \caption{Performance of progress-enhanced policy learning. The metrics are the task and goal completion percentage, with the path-weighted scores given in parentheses.}
    \label{tab:application_progress_estimation}
    \begin{adjustbox}{width=\linewidth}
        \begin{tabular}{lcccc}
            & \multicolumn{2}{c}{\textit{Valid\_Seen}} & \multicolumn{2}{c}{\textit{Valid\_Unseen}} \\
            \toprule
            Model & Task & Goal-Cond & Task & Goal-Cond \\
            \midrule
            Seq2Seq\textsuperscript{*}                                        & 2.4 (1.1) & 9.4 (5.7) & 0.1 (0.0) & 6.8 (4.7)  \\
            \hspace{1pt}+ PM TimeBased\textsuperscript{*}            & 2.1 (1.1) & 8.7 (5.6) & 0.0 (0.0) & 6.9 (5.0)  \\
            \hspace{1pt}+ PM \method           & \B{3.4} (\B{2.4}) & \B{10.9} (\B{7.9}) & \B{0.2} (\B{0.1}) & \B{7.3} (\B{5.1})  \\
            \bottomrule
        \end{tabular}
    \end{adjustbox}
    \footnotesize{Results marked with * are reported from the original ALFRED paper.}
    \vskip -0.15in
\end{table}

\begin{table*}[t!]
    \centering
    \setlength{\tabcolsep}{12pt}
    \caption{Performance on the reward modeling task. We perform online reinforcement learning based on the pretrained SPRINT model on the \textit{EVAL}\(_{\textit{INSTRUCT}}\) benchmark, and report the step and task completion rate.}
    \label{tab:application_online_rl}
    \begin{adjustbox}{width=\linewidth}
        \begin{tabular}{lcccccccccc}
          \toprule
          & \multicolumn{2}{c}{2 steps} & \multicolumn{2}{c}{3 steps} & \multicolumn{2}{c}{4 steps} & \multicolumn{2}{c}{5 steps} & \multicolumn{2}{c}{overall} \\
          \cmidrule(lr){2-3} \cmidrule(lr){4-5} \cmidrule(lr){6-7} \cmidrule(lr){8-9} \cmidrule(lr){10-11}
          & Step & Task & Step & Task & Step & Task & Step & Task & Step & Task \\
          \midrule
            Pretrained SPRINT                      & 1.35 & 0.55 & 1.90 & 0.50 & 2.25 & 0.40 & 3.31 & 0.23 & 2.20 & 0.42 \\
          \midrule
            Step-Completion Based Reward           & 1.30 & 0.55 & 1.85 & 0.50 & 2.15 & 0.40 & 2.77 & 0.23 & 2.02 & 0.42 \\
            Qwen2.5-VL-Instruct                    & 1.55 & 0.65 & 1.55 & 0.35 & 1.75 & 0.25 & 2.61 & 0.23 & 1.87 & 0.37 \\
            \method                        & \textbf{1.65} & \textbf{0.70} & \textbf{2.20} & \textbf{0.60} & \textbf{2.55} & \textbf{0.45} & \textbf{3.38} & 0.23 & \textbf{2.45} & \textbf{0.50} \\
          \bottomrule
        \end{tabular}
    \end{adjustbox}
\end{table*}

\begin{figure*}[h!]
    \centering
    \includegraphics[width=0.95\linewidth]{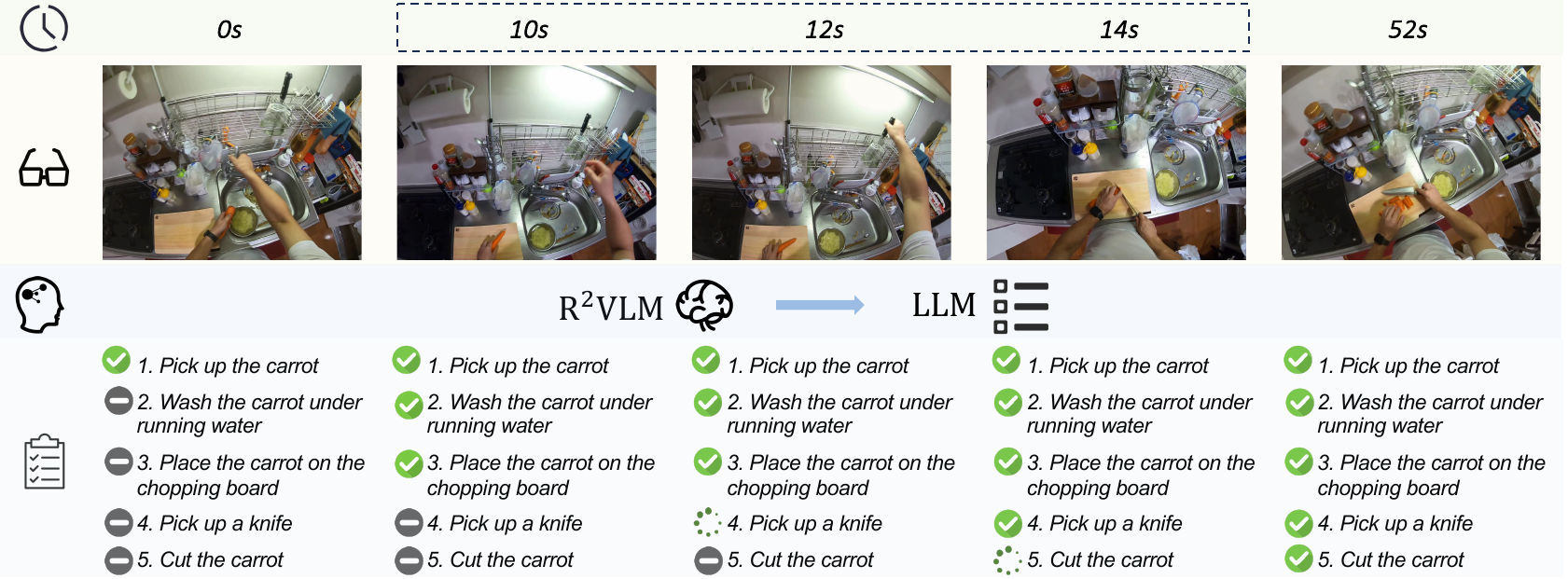}
    \caption{Demonstration of \method as a proactive assistant. The monitoring task is ``cutting a carrot on a chopping board''.}
    \label{fig:proactive_assistance}
\end{figure*}

\noindent\textbf{Reward Modeling for Online Reinforcement Learning.}
Another key application of our model is to provide general and dense reward signals for reinforcement learning of agent policy models~\cite{Cheng_2024_CVPR,cheng2024videgothink}. As the progress indicates the advancement of a task, the progress increment after each action can be naturally interpreted as its reward. We validate this idea in the AI2-THOR 2.0 simulator using the Alfred-RL implementation from SPRINT~\cite{zhang2024sprint}, which implements online IQL~\cite{kostrikov2021offline} for fine-tuning policy models. The original SPRINT uses a sparse step-completion reward (1 only when a step is finished). We adopt this as our baseline and compare it with dense rewards derived from \method and the original Qwen2.5-VL-Instruct-7B.

We fine-tune the pretrained SPRINT policy model for 50k online RL steps on the $\textit{EVAL}_{\textit{INSTRUCT}}$ benchmark, which contains long-horizon, unseen instruction-following tasks. Following SPRINT, we separately train policies on scenarios with 2–5 steps. As illustrated in~\cref{tab:application_online_rl}, our model provides the most effective reward signals. Agents trained with \method rewards consistently improve across all task lengths, whereas sparse or Qwen2.5-VL-Instruct rewards often reduce the performance. In~\cref{fig:reward_modeling}, we further compare the agent's execution trajectories and the corresponding rewards from \method before and after the RL stage. This demonstrates that \method can provide well-aligned rewards and can effectively guide the model to produce correct action sequences.

\noindent\textbf{Step-wise Proactive Assistance.}
Our model enables proactive assistance by continuously receiving streaming video and periodically reasoning over new snippets to update step-level completion status. This information can help human operators or autonomous agents understand the execution status and make informed decisions.
We conduct qualitative experiments on Ego4D, where \method processes 2-second video snippets and continually updates the CoT recording step statuses. A lightweight LLM (Qwen2.5-7B-Instruct) then be used to extract structured lists of completed, in-progress, and pending steps from the updated CoT. As shown in~\cref{fig:proactive_assistance}, the system accurately detects key transitions (at $10 \rightarrow 12 \rightarrow 14$s) and rapidly updates step states when relevant visual cues appear. This enables continuous, near real-time tracking of task execution, supporting more intelligent and collaborative guidance in contexts such as education, accessibility, and smart fitness.

\section{Conclusion}

We propose \method, a recurrent reasoning vision-language model for long-horizon task progress estimation. 
By combining an evolving CoT with iterative reasoning over local video snippets, \method maintains global context while avoiding the computation overhead of processing full video trajectories. 
Experiments on ALFRED and Ego4D datasets show that this design yields accurate, generalizable, and robust progress estimation performance. Moreover, the progress signals benefit diverse downstream tasks, including progress-enhanced policy learning, reward modeling for agent reinforcement learning, and step-wise proactive assistance. Overall,  offers an efficient and practical solution for progress estimation tasks in embodied AI.

{
    \small
    \bibliographystyle{ieeenat_fullname}
    \bibliography{reference}
}

\appendix
\clearpage
\setcounter{page}{1}
\maketitlesupplementary

\section{Datasets \& Benchmarks}
\subsection{Dataset Information}
\label{sec:appendix_dataset_statistics}

We leverage expert demonstrations from two datasets: ALFRED~\cite{shridhar2020alfred} (simulation) and Ego4D~\cite{grauman2022ego4d} (real world). The ALFRED dataset consists of 25,743 human-annotated high-level tasks and 8,055 corresponding videos, collected within the AI2-THOR 2.0~\cite{kolve2017ai2thor} simulator. Each video is annotated with start and end timestamps for step-level decompositions. Ego4D is a well-known real-world egocentric dataset. We use the goal-step annotated subset~\cite{song2023ego4dgoalstep}, which provides start and end time annotations for 48,000 tasks. In addition, Ego4D offers narration data that specifies the start times of human actions. We define the end time of each step as the start time of the subsequent one. Thus, data from both datasets can be jointly represented as:
$\{\mathcal{T}, \mathcal{E}, (s_i, t^s_i, t^e_i), i=1 \dots n\}$
where $\mathcal{T}$ denotes the task description, $\mathcal{E}$ the video expert demonstration, and $n$ represents the total number of steps in the task. Each step $s_i$ is associated with a start time $t^s_i$ and an end time $t^e_i$.

The frame rate of ALFRED video demonstrations is 5 FPS. We downsample the videos to 1 FPS and construct video snippets at 4-second intervals. While the frame rate of Ego4D goal-step subset is 30 FPS, we downsample it to 2 FPS and segment at 2-second intervals. Both datasets provide annotations for task description, step descriptions, and their corresponding start and end times. 
In the main paper, we detail the pipeline for constructing the incremental video data and generating the cumulative Chain-of-Thought in~\cref{sec:dataset}. The default large models used in our data construction process are Qwen2.5-VL-72B-Instruct and Qwen2.5-72B-Instruct. The statistical information of the generated dataset is provided in~\cref{tab:appendix_dataset_information}.

\begin{table}[h!]
    \centering
    \caption{Detailed statistics information of two generated datasets.}
    \label{tab:appendix_dataset_information}
    \begin{adjustbox}{width=0.98\linewidth}
        \begin{tabular}{lcc}
          \toprule
          & Alfred Dataset & Ego4D Dataset \\
          \midrule
            Dataset size                   & 11499  & 13965  \\
            Total number of turns          & 124821 & 127694 \\
            Average turns per trajectory   & 10.85  & 9.14   \\
            Average video length (seconds) & 47.58  & 22.20  \\
            Frame resolution (pixels)      & 300 $\times$ 300 & 320 $\times$ 240 \\ 
          \bottomrule
        \end{tabular}
    \end{adjustbox}
\end{table}

We construct the benchmarks using the same procedure under unseen scenarios. Three volunteers manually filtered the data. In ALFRED, 669 out of 722 automatically annotated samples were retained, resulting in a 93\% retention rate. In Ego4D, 529 out of 718 samples were kept, yielding a 74\% retention rate.
The Ego4D environment is inherently more complex, and the large model may not have taken the entire environment into full account during data generation, leading to slightly lower data quality than in ALFRED. Even so, our dataset still maintains a 74\% retention rate, indicating satisfactory data quality.

\subsection{Details of Generating Distractor  Task Descriptions}
\label{sec:appendix_distractor_detail}

Introducing distractor descriptions during training prevents the model from relying on visual variations as shortcuts for inferring task progress. It instead forces the model to jointly reason over task descriptions and visual inputs, improving robustness and applicability in complex downstream environments filled with distractions.

Here, we detailed our strategy for generating distractor task descriptions. The crucial point is that the ground-truth progress corresponding to the generated distractor description must be explicitly determinable. This means that such descriptions cannot be produced by directly prompting a large model, as the resulting progress values would be undefined. To address this, we develop a constrained prompt strategy. During generation, the VLM is required to produce distractor descriptions that overlap with the initial $n_r$ steps while remaining unrelated to all later steps. Additionally, we constrain the generated descriptions to contain the same number of sub-steps as the original. This design enables an accurate mapping of distractor descriptions to corresponding progress in the expert demonstration. When the expert video is executing the first $n_r$ steps, the progress corresponding to the distractor description increases normally. At the end of step $n_r$, the progress reaches its maximum value of $100 \cdot \frac{n_r}{n}$. After that point, the progress associated with the distracting description stays constant. The complete formulation is:
\begin{align}
    \textstyle p_t \;=\; 100 \cdot \min\!\left( \frac{n_r}{n}, \; \frac{k}{n} + \frac{1}{n}\cdot \frac{t - t_{k+1}^s}{t_{k+1}^e - t_{k+1}^s} \right).
\end{align}

Furthermore, to enhance the quality and difficulty of the distracting descriptions, we extract existing objects and entities from the environment as candidate elements during generation. The prompt used for generating the distracting descriptions is shown in~\cref{sec:appendix_distractor_generation_prompt}.

\section{Training \& Evaluation}
\subsection{Training Details}
\label{sec:appendix_training_details}

We leverage %
\textit{LLaMA-Factory} framework for supervised finetuning, and %
\textit{Verl} for RL. To adapt to our data format, we modify the multi-turn rollout algorithm in the Verl framework by removing the history contexts of previous turns and adding the cumulative chain of thought. We also add the multi-modal support in the multi-turn rollout code. The source code and training scripts will be made available.

Both the SFT and RL stages are trained using 8 GPUs, with full-parameter fine-tuning. The learning rate is set to 1e-5, and the maximum number of output tokens is 1024. For reinforcement learning, we employ the PPO algorithm with the KL coefficient set to 0. 
For the ALFRED dataset, the SFT stage is conducted with a batch size of 64 for 8 epochs (15k steps). We use the checkpoint optimized for 5,000 steps as the cold-start model and continue training it with reinforcement learning. In the RL stage, we select longer multi-turn data with at least 5 turns to enhance the model’s ability in long-horizon reasoning. The batch size for RL is 32, trained for 1.5 epochs. The RL-Zero model is trained for 3.5 epochs, during which the overall reward is observed to have converged.
On the Ego4D dataset, the SFT stage is trained with a batch size of 64 for 8 epochs (16k steps). The checkpoint obtained after 10k optimization steps during SFT is used to initialize the subsequent RL training. During the RL stage, we select challenging multi-turn interaction data with more than 10 rounds. The RL batch size is also set to 32, trained for 4 epochs. The RL-Zero model is trained for 8 epochs.

We use DeepSpeed for SFT training. The ZeRO-3 offload mode is enabled only when training models with full video inputs, whereas ZeRO-2 optimization is used for the remaining experiments. To ensure fairness, all ablation studies are conducted under identical training settings. Further training parameters are available in our released scripts.

\subsection{Baseline Model Information}
\label{sec:appendix_baseline_model_information}

\begin{table}[h!]
    \centering
    \caption{Detailed number of input frames of each baseline model.}
    \label{tab:appendix_baseline_model_information}
    \begin{adjustbox}{width=\linewidth}
        \begin{tabular}{lcc}
            \toprule
            Model Name & Input Frames (Alfred) & Input Frames (Ego4D)    \\
            \midrule
            \rowcolor{gray!10}
            \multicolumn{3}{c}{\textbf{\textit{API-Based Methods}}} \\ 
            \midrule
            GPT-5~\cite{openai2025gpt5}                 & 32  & 32  \\
            Gemini-2.5-Pro~\cite{comanici2025gemini}    & 16  & 16  \\
            Qwen3-VL-Plus~\cite{qwen2025qwen3vlmax}     & 32  & 32  \\
            \midrule
            \rowcolor{gray!10}
            \multicolumn{3}{c}{\textbf{\textit{Open-Source Models}}} \\ 
            \midrule
            MiniCPM-V-2.6~\cite{yao2024minicpm}         & 8   & 8   \\
            GLM-4.1V-Thinking~\cite{hong2025glm}        & 32  & 32  \\
            InternVL3~\cite{zhu2025internvl3}           & 32  & 8   \\
            Qwen2.5-VL~\cite{bai2025qwen25vl}           & 32  & 32  \\
            \bottomrule
        \end{tabular}
    \end{adjustbox}
\end{table}

The pretrained weights of all open-sourced models can be downloaded from HuggingFace. For LIV~\cite{ma2023liv}, we use the model to compute the similarity between the textual task description and each frame, and normalize the values to 0-100 to represent the progress percentage. GVL~\cite{ma2024vision} and ROVER~\cite{schroeder2025rover} are methods based on in-context learning. 
For a fair comparison, we fine-tune GVL on our generated dataset based on Qwen2.5-VL-7B, resulting in GVL-SFT, as reported in~\cref{tab:main_results}.
Due to the difficulty of constructing data for ROVER's recursive reasoning framework, we follow the original setting and directly prompt VLM to produce the results. 
Since the Gemini-1.5-Pro model employed in the original implementation was officially discontinued on September 24, 2025, we use Qwen2.5-VL-72B as a replacement in our experiments.

\subsection{Evaluation Metric Formulations}
\label{sec:appendix_evaluation_metric_formulations}

We compute the average metrics by performing an average at each iteration, which results in trajectories with more iterations receiving larger weights.
For evaluation, we deploy the model on a single GPU using vLLM and use multithreading to access the server for accelerated inference. When measuring the time cost of different video input formats in~\cref{sec:ablation_studies}, we use the same machine and environment and fix the number of threads to 16. The evaluation scripts will also be released.

Same as in \cref{sec:RL-training}, at iteration $t$, let $p_t$ and $p_t^{\text{gt}}$ denote the predicted and ground-truth progress, respectively, while $n^{\text{gt}}$ and $m_t^{\text{gt}}$ represent the ground-truth total number of subtasks and the number of completed subtasks. We present the formulations of the four metrics in the following.

\begin{align}
    p_{\text{mae}} &= \lvert p_t - p_t^{\text{gt}}\rvert, \\
    \Delta p_{\text{mae}} &= \lvert (p_t - p_{t-1}) - (p_t^{\text{gt}} - p_{t-1}^{\text{gt}})\rvert, \\
    bin &=
    \begin{cases}
        1.0,  & \text{if } \frac{m_t^\text{gt}}{n^\text{gt}} \leq \frac{p_t}{100} <\frac{ (m_t^\text{gt} + 1)}{n^\text{gt}};  \\
        0.0.  & \text{otherwise},
    \end{cases}  \\
    acc &=
    \begin{cases}
      1.0, & \text{if } p_t = 100 \land p_t^\text{gt} = 100;  \\
      1.0, & \text{if } p_t \neq 100 \land p_t^\text{gt} \neq 100;  \\
      0.0, & \text{otherwise}.
    \end{cases}
\end{align}

Since iterations with $p_t^{\text{gt}} < 100$ account for the majority of each trajectory, we calculate the proportions of the two iteration types ($p_t^{\text{gt}} = 100$ vs. $p_t^{\text{gt}} < 100$) in each benchmark. We then weight the accuracy score using the inverse of their proportions to reduce distribution bias and obtain a more accurate evaluation of the model’s task-completion judgment.

Each of the four metrics, as detailed in~\cref{sec:experiment_setups}, provides an objective and unambiguous assessment of the model’s capability on downstream tasks. $\text{R}^2$VLM surpasses all baseline models across all four metrics and demonstrates strong performance on the three downstream applications, including progress-enhanced embodied policy learning, reward modeling for online reinforcement learning, and step-wise proactive assistance, which demonstrates its considerable application potential.

\section{Cross-Domain Generalization Analysis}
\label{sec:appendix_crossdomain}
We evaluate the cross-domain generalization of $\text{R}^2$VLM by training on one dataset and testing on another. We also report results obtained by training on the combined datasets, as summarized in~\cref{tab:model_generalization}. The results show that models trained with cross-domain data consistently outperform the zero-shot baseline. Moreover, training on both datasets further improves performance on most metrics, with particularly notable gains on Ego4D. The strong cross-domain performance of $\text{R}^2$VLM suggests that the model acquires genuine reasoning and analytical capabilities, instead of simply memorizing dataset-specific patterns.

\begin{table}[h!]
    \centering
    \caption{Cross-domain generalization evaluation of $\text{R}^2$VLM.}
    \label{tab:model_generalization}
    \begin{adjustbox}{width=\linewidth}
        \begin{tabular}{lcccc|cccc}
          \toprule
            & \multicolumn{4}{c}{Alfred} & \multicolumn{4}{c}{Ego4d} \\
            Train Dataset & $p_\text{mae}\downarrow$ & $\Delta p_\text{mae}\downarrow$ & $bin\uparrow$ & $acc\uparrow$ & $p_\text{mae}\downarrow$ & $\Delta p_\text{mae}\downarrow$ & $bin\uparrow$ & $acc\uparrow$ \\
          \midrule
            None               & 27.87           & 10.61           & 0.295            & 0.494            & 28.32            & 10.11           & 0.206            & 0.485          \\
            Alfred             & \textbf{2.77}   &\underline{2.27} &\textbf{0.876}    &\underline{0.961} & 25.03            & 5.87            & 0.252            & 0.539          \\
            Ego4D              & 19.64           & 6.56            & 0.336            & 0.611            &\underline{23.03} &\underline{3.98} &\underline{0.266} &\underline{0.685}  \\
            Alfred + Ego4D     &\underline{3.02} &\textbf{2.26}    &\underline{0.862} &\textbf{0.968}    &\textbf{22.68}    &\textbf{3.67}    &\textbf{0.268}    &\textbf{0.714}\\
          \bottomrule
        \end{tabular}
    \end{adjustbox}
\end{table}

\section{More Details of Three Downstream Application Experiments}

Here in this section, we provide more details and analysis of the three downstream application experiments in~\cref{sec:application}.

\subsection{Progress-enhanced Embodied Policy Learning}

This experiment originates from the Alfred paper, where the authors argue that tracking progress toward the goal helps agents assess the utility of intermediate states and ultimately improve task performance. To this end, they added an auxiliary head to the Seq2Seq model to predict task progress and incorporated the prediction loss as additional supervision during training.

One setting used in Alfred defines progress based on execution time, assuming that progress increases linearly throughout the task. This definition is clearly suboptimal, as it fails to capture the inherent heterogeneity of task steps in terms of duration and action density. Some critical steps can significantly accelerate task advancement within a short period, whereas others may take considerable time while contributing only marginally to actual progress.

As most datasets in the AI2-THOR simulator do not provide explicit annotations of task progress, $\text{R}^2$VLM can serve as a general-purpose progress estimator. By defining progress based on the proportion of completed steps, $\text{R}^2$VLM offers a more faithful estimate of task advancement. In this experiment, we compare the effectiveness of progress estimates provided by $\text{R}^2$VLM with those based on execution time in assisting policy learning for embodied agents. Specifically, the time-based progress can be directly computed, whereas the estimated progress is obtained by using $\text{R}^2$VLM to annotate the entire training set. The results in~\cref{tab:application_progress_estimation} clearly demonstrate the superiority of the progress estimates produced by our model, not only highlighting their practical value in assisting agent policy learning but also validating the effectiveness of our proposed progress definition strategy.

\subsection{Reward Modeling for Online Reinforcement Learning}

This experiment validates the effectiveness of the proposed progress estimation model for applying in online reinforcement learning tasks. Given that progress can be interpreted as a general value estimate of the agent's state~\cite{ma2024vision,schroeder2025rover}, the change in progress thus corresponds to the immediate reward of the executed action, as derived from the Bellman Equation with $\gamma = 1$. 

In this experiment, we leverage the Alfred-RL~\cite{zhang2024sprint} source code to conduct online reinforcement learning on top of the pretrained SPRINT~\cite{zhang2024sprint} model. The SPRINT model is pretrained on LLM-composed task sequences and possesses basic capability for executing multi-step tasks. We further perform reinforcement learning on tasks from the $\textit{EVAL}_{\textit{INSTRUCT}}$ benchmark, which contains complex, previously unseen tasks with longer action sequences (up to 5 steps). We compare the step–completion–based reward used in the original paper with the rewards generated by Qwen2.5-VL-Instruct-7B and our $\text{R}^2$VLM. The results in~\cref{tab:application_online_rl} demonstrate the strong reward-modeling capability of our model, which significantly surpasses the two baselines.

Note that $\text{R}^2$VLM is pretrained using four input frames. However, in the online RL stage, the simulator only provides one observation image after each action. In our implementation, we duplicate the pre-action image twice and the post-action image twice to form the input. This design is mainly to match the input format used during $\text{R}^2$VLM pretraining. We believe that training a single-frame version of $\text{R}^2$VLM specifically for this task could further improve its effectiveness. \cref{fig:reward_modeling} visualizes the trajectories and corresponding rewards of the policy model executing on the same task before and after reinforcement learning. As shown, the rewards provided by $\text{R}^2$VLM effectively guide the model toward executing the correct actions. For example, the pretrained SPRINT model fails to pick up the red pen specified in the task description and instead grabs the keychain. Our model correctly identifies this mistake and assigns no reward. In contrast, when $\text{R}^2$VLM observes the correct pickup of the red pen, it provides a reward of 0.20. This illustrates the model's reward modeling capability.

We argue that training embodied agents with reinforcement learning is a highly challenging process, especially when the task consists of multiple sequential steps (such as the five-step long-horizon tasks in this experiment). In such cases, if the reward signals are too sparse or inaccurate, the training process can easily become unstable or even collapse. We present a method that utilizes VLMs to deliver dense, precise, and general rewards for embodied tasks, which effectively addresses the challenge above and offers a new solution for reward modeling and policy learning in complex scenarios. This feasibility is also supported by our input formulation that integrates short video snippets with historical CoT, enabling the model to quickly provide rewards based on the agent's post-action observations, thus serving as an online reward provider. 

\subsection{Step-wise Proactive Assistance}

This experiment investigates the effectiveness of our progress estimation model, which is designed to process streamed video, in providing proactive assistance. Our model performs inference every two seconds based on the input streamed video. It integrates the historical CoT with the newly observed video snippet to conduct recurrent reasoning and update the CoT. The updated CoT, which records the step completion status, is then passed to an LLM to produce a structured summary of the completed, in-progress, and pending steps for users. This is a straightforward task, and we use Qwen2.5-Instruct-7B in practice. Such step-wise progress feedback helps both agents and human operators maintain a clearer understanding of the current task execution state and assists better decision-making in complex tasks. This capability has promising applications in areas such as education and healthcare, and can be deployed on intelligent wearable devices.

\clearpage
\onecolumn
\section{Visualization of More Results}

\subsection{Step-by-Step Progress Estimation}

\begin{figure}[htbp]
  \centering
  \begin{subfigure}[b]{0.482\textwidth}
    \centering
    \includegraphics[width=\textwidth]{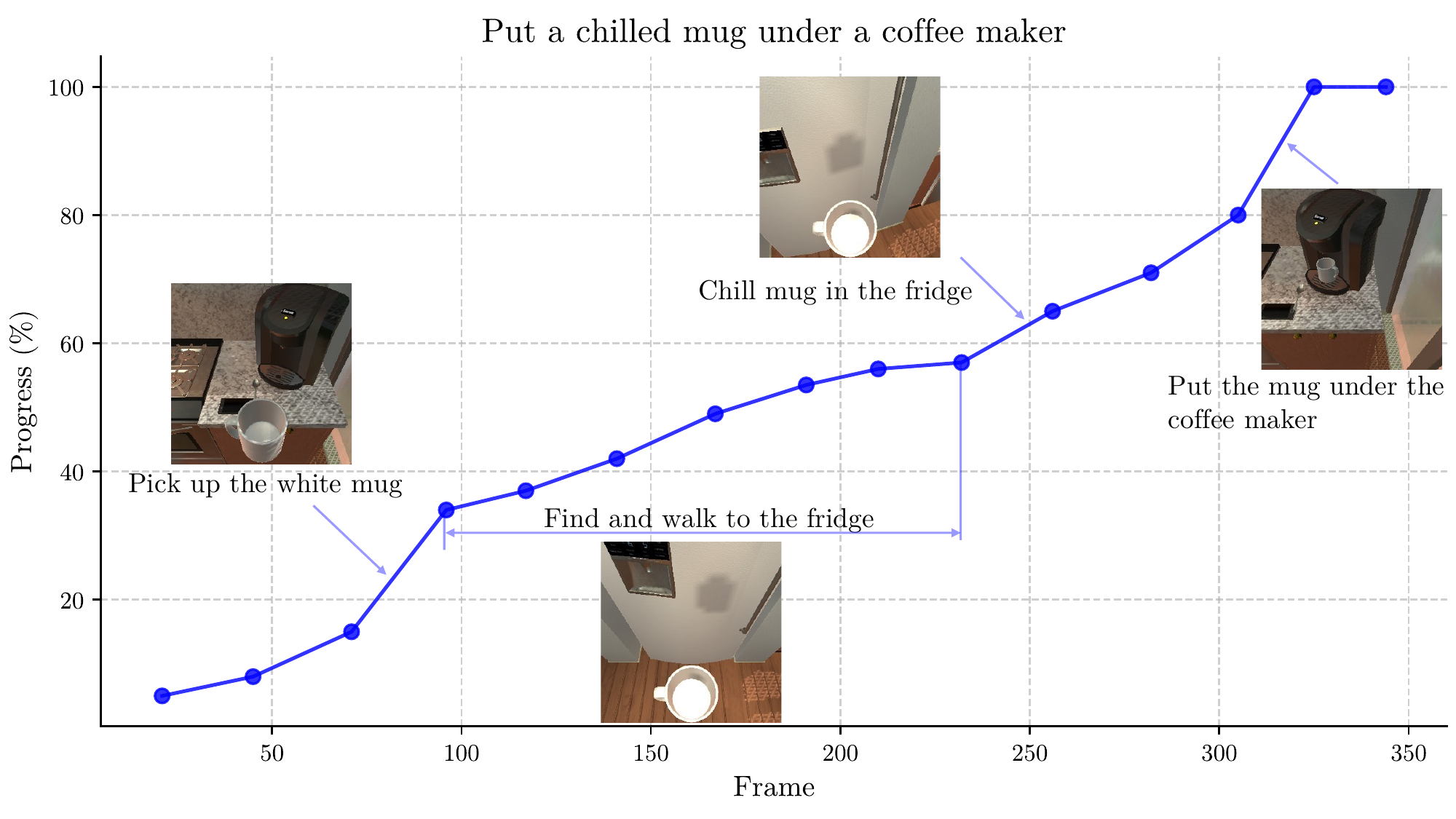}
  \end{subfigure}
  \hfill
  \begin{subfigure}[b]{0.482\textwidth}
    \centering
    \includegraphics[width=\textwidth]{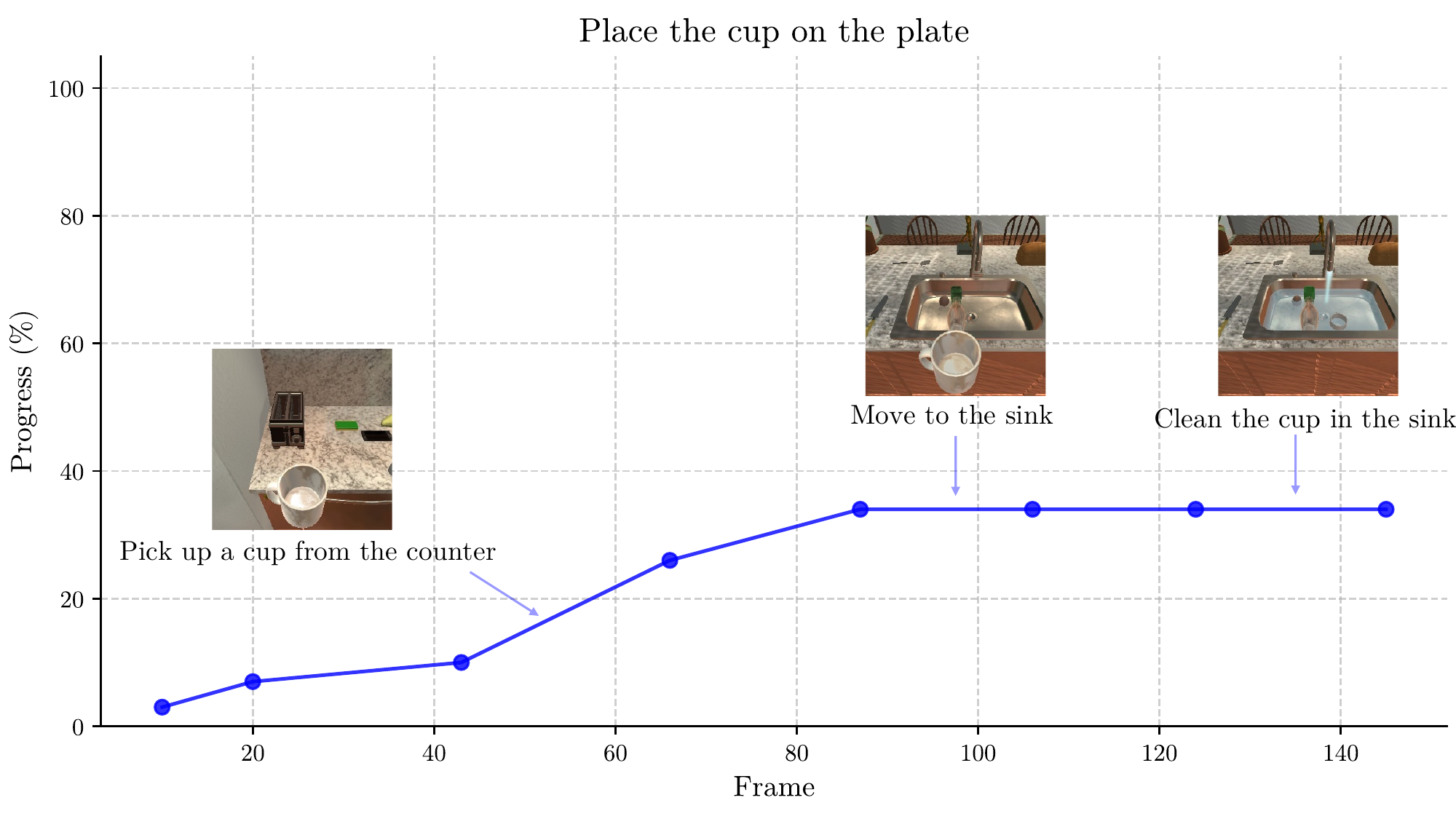}
  \end{subfigure}

  \begin{subfigure}[b]{0.482\textwidth}
    \centering
    \includegraphics[width=\textwidth]{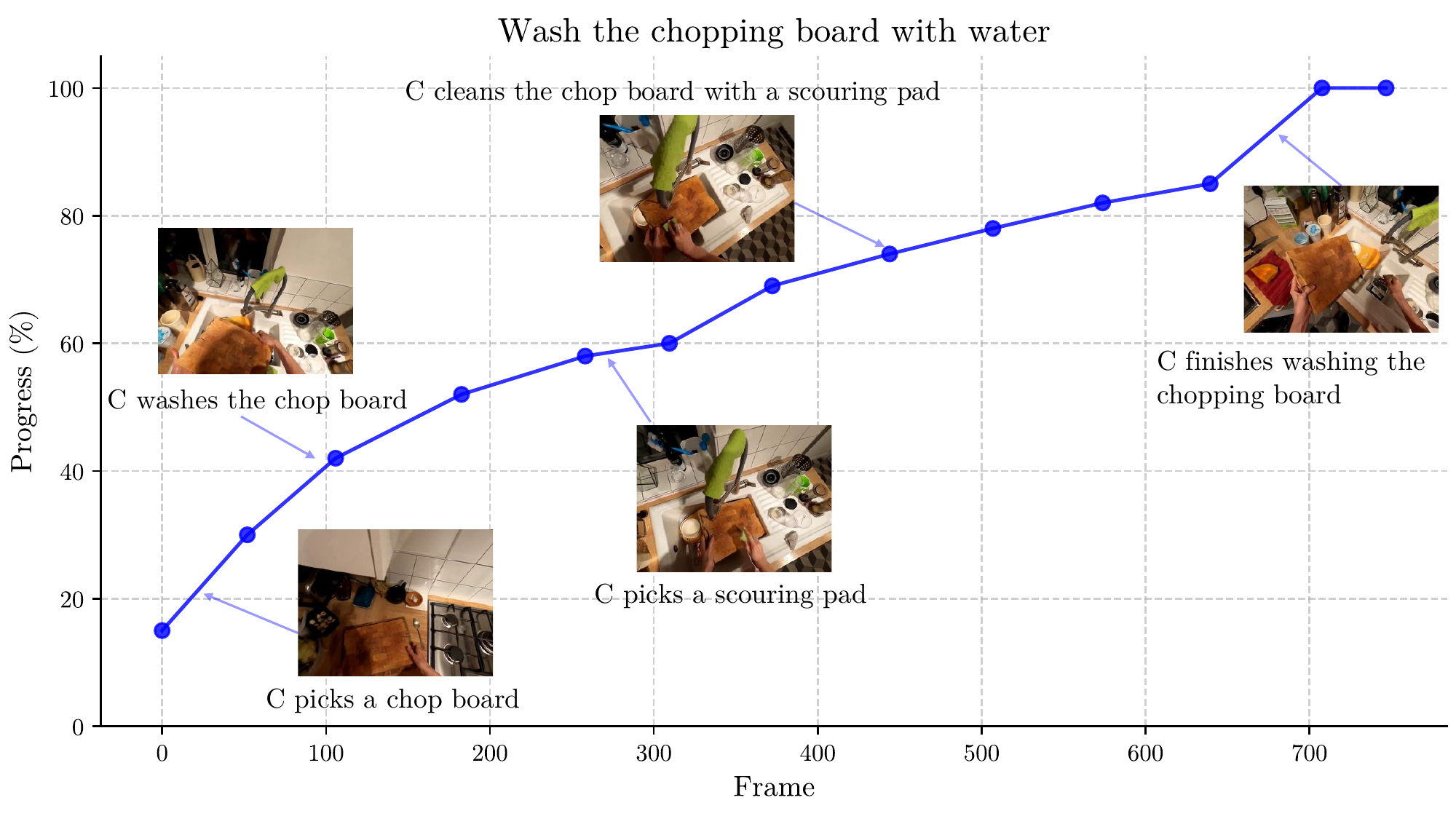}
  \end{subfigure}
  \hfill
  \begin{subfigure}[b]{0.482\textwidth}
    \centering
    \includegraphics[width=\textwidth]{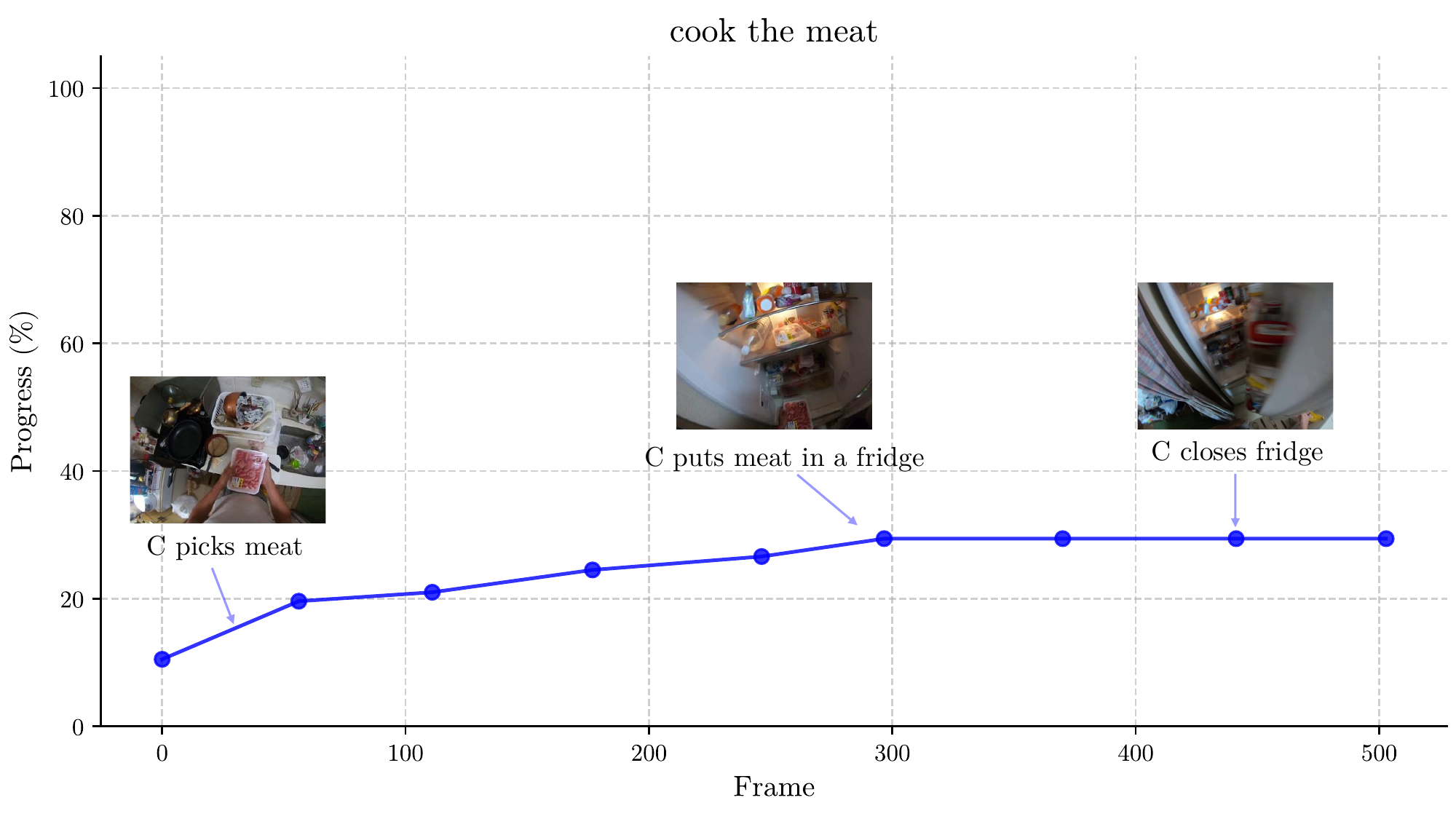}
  \end{subfigure}

  \caption{We visualize the model’s progress estimation on four tasks, where key steps trigger sharp progress increases, and irrelevant steps yield zero progress increments.}
  \label{fig:appendix_progress_estimation_demo}
\end{figure}

We visualize the step-by-step progress estimations for four data points in~\cref{fig:appendix_progress_estimation_demo}. The upper plots correspond to the Alfred benchmark, and the lower ones to the Ego4D benchmark. The left column shows visualizations of original descriptions, while the right column displays the progress of distractor descriptions. As shown in the figures, the occurrence of key steps triggers sharp increases in progress, whereas irrelevant steps produce no progress increments. The progress estimated for distractor description increases only in the early stage when relevant steps appear in the expert trajectory, and remains unchanged thereafter.

\subsection{Dynamic CoT Revision}

\begin{figure}[h!]
    \centering
    \includegraphics[width=0.83\linewidth]{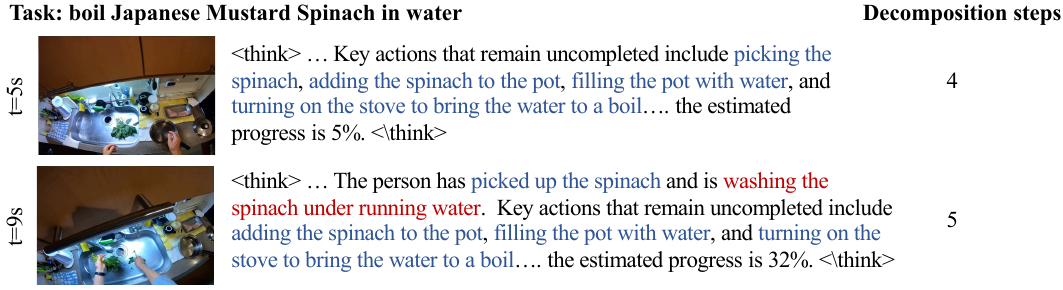}
    \caption{A demo of \method dynamically revising the CoT.}
    \label{fig:failure_case}
\end{figure}

The CoT revision mainly occurs when the initial task decomposition conflicts with the actual execution in the video. As illustrated in~\cref{fig:failure_case}, at $t=5s$ the model generates an initial CoT from prior knowledge, while at $t=9s$ it observes the spinach being washed and thus revises the CoT accordingly to match the visual evidence.

\clearpage

\section{Prompts}

\subsection{Model System Prompt}
\label{sec:appendix_model_system_prompt}

\begin{tcolorbox}[
    breakable, 
    colback=MyBlue!5!white, 
    colframe=MyBlue!90!black, 
    title={The System Prompt Defines the Task and Reasoning Strategy.}
]
\begin{lstlisting}[basicstyle=\ttfamily\small, breaklines=true]
You are a professional assistant skilled in understanding temporal videos and estimating task progress.
You will be given a history chain of thought (CoT) summarizing completed and remaining steps, along with incremental egocentric video frames.
Your task is to:
1. Analyze the new frames to detect any completed actions or state changes.
2. Update the list of completed steps and revise the remaining steps accordingly. (Merge, split, or reorder steps if necessary)
3. Estimate the overall task completion progress as a percentage, based on the proportion of completed steps made toward the full task. (eg. finish 1 of 3 tasks, progress is 33~66)
Use the following output format:
<think>
[Updated chain of thought: clearly list revised completed steps and remaining steps.]
</think>
<answer>
[Task completion percentage based on the proportion of completed steps, an integer between 0 and 100]
</answer>

History CoT: {history_cot}
Images: <image><image><image><image>
Question: {question}
Output: 
\end{lstlisting}
\end{tcolorbox}

\subsection{Distractor Task Description Generation Prompt}
\label{sec:appendix_distractor_generation_prompt}

\begin{tcolorbox}[
    breakable, 
    colback=MyBlue!5!white, 
    colframe=MyBlue!90!black, 
    title={The Prompt to Generate Distractor Task Description.}
]
\begin{lstlisting}[basicstyle=\ttfamily\small, breaklines=true]
You are given a task description, a sequence of initial steps, and a set of available objects in the environment.

Your goal is to generate a distractor task description that meets the following conditions:
1. The distractor task description must share the same initial steps.
2. The distractor task must ultimately be different from the original task.
3. The distractor should have the same number of substeps as the original description.
4. The distractor task should be constructed using one or more objects from the provided object list.
5. The distractor should be natural and plausible in the given environment.

Only output the distractor task description, without any additional explanations or context.

Example1:
Input:
- Original task description: Put a wine glass in the refrigerator.
- Initial steps: 
    Turn around and go to the kitchen counter, to the right of the sink.
- Objects: Apple, SaltShaker, Mug, Pot, Cup, Egg, SoapBottle, Bread, DishSponge, Plate

Output:
Place the apple from the counter onto the plate.
---

Example2:
Input:
- Original task description: Put heated egg on kitchen table.
- Initial steps: 
    walk to face sink
    pick up egg from sink
    walk to face microwave
- Objects: Bowl, Pan, Lettuce, PepperShaker, Cup, Egg, SoapBottle, Bread, DishSponge, ButterKnife, Apple, SaltShaker, Mug, Pot, Potato, Knife, Plate, Spoon, Fork, Tomato, Spatula

Output:
Put the egg into the bowl and pick up the bowl.

Now it's your turn:

Original task description: {task_description}
- Initial steps: 
{narrations}
- Objects: {objects}

Output: 
\end{lstlisting}
\end{tcolorbox}

\subsection{Chain of Thought Generation Prompt}
\label{sec:appendix_cot_generation_prompt}

\begin{tcolorbox}[
    breakable, 
    colback=MyBlue!5!white, 
    colframe=MyBlue!90!black, 
    title={The Prompt to Generate Chain of Thought.}
]
\begin{lstlisting}[basicstyle=\ttfamily\small, breaklines=true]
You will be given:
1. A **video** and a question about the **progress** of a **task** shown in the video.
2. A list of **completed narrations** and **uncompleted narrations**.

Your job is to generate a human-style chain of thought describing the task's progress, based on your understanding of how the task is completed and the provided completed and uncompleted narrations.

You must:
- Analyze what important steps have been completed and how they contribute to the overall task progress.
- Analyze what key steps remain uncompleted and how crucial they are for finishing the task.
- Give a final estimated progress percentage **based on the proportion of completed steps to the total number of steps.**
- Treat the narrations only as background guidance to simulate watching the video, and base your reasoning on the video content. Do not use any language implying reliance on the narrations or mentioning the phrase "completed narrations" or "uncompleted narrations."
---

Now begin your analysis:

Task:  
{task_description}

Completed Narrations:
{completed_narrations}

Uncompleted Narrations:
{uncompleted_narrations}

Progress:
{progress}

Output:

Your output should follow this format:
<think>
To determine the progress of the task "{task_description_lower}", let's analyze the video:
[Explain which actions were completed and their significance in moving toward the goal.]
[Explain which key actions remain uncompleted and why they are crucial.]
[Conclude with a reasonable progress estimate **based on the proportion of completed steps to the total number of steps**.]
</think>
<answer>
You must strictly output the same final progress percentage I provided above: {progress}
</answer>
\end{lstlisting}
\end{tcolorbox}

\begin{tcolorbox}[
    breakable, 
    colback=MyBlue!5!white, 
    colframe=MyBlue!90!black, 
    title={Examples of Generated Chain of Thought.}
]
\begin{lstlisting}[basicstyle=\ttfamily\small, breaklines=true]
<think>
To determine the progress of the task "put a clean potato slice on the counter," let's analyze the video:

The robot has already turned around and faced the counter, picked up the knife, crossed the room to face the stove, sliced the potato, placed the knife back on the counter, and taken a piece of potato. These actions indicate significant progress towards preparing the potato slice. The slicing step is crucial as it directly contributes to creating the potato slice needed for the task.

However, several key steps remain uncompleted. The robot needs to cross the room to the sink, wash the potato slice under running water, and then place the clean slice on the counter. Washing the potato is essential to ensure it is clean before placing it on the counter, making these remaining steps critical for completing the task.

Considering the proportion of completed steps to the total number of steps, the progress appears to be substantial but not yet complete. The slicing part is done, but the cleaning and final placement are still pending. Based on this reasoning, a reasonable estimate of the progress would be around 65%

</think>
<answer>
67
</answer>
\end{lstlisting}
\end{tcolorbox}

\end{document}